\documentclass{article}

\usepackage{microtype}
\usepackage{graphicx}
\usepackage{subfigure}
\usepackage{booktabs} 
\usepackage[mathscr]{eucal}
\usepackage{algorithm}
\usepackage{algorithmic}
\usepackage{wrapfig}
\usepackage{amsmath}
\usepackage[font=small,labelfont=bf]{caption}

\usepackage{hyperref}

\usepackage[accepted]{icml2019}

\icmltitlerunning{Interpretable Multi-Modal Hate Speech Detection}

\begin{document}

\twocolumn[
\icmltitle{Interpretable Multi-Modal Hate Speech Detection}




\begin{icmlauthorlist}
\icmlauthor{Prashanth Vijayaraghavan}{mit}
\icmlauthor{Hugo Larochelle}{goo}
\icmlauthor{Deb Roy}{mit}
\end{icmlauthorlist}

\icmlaffiliation{goo}{Google Brain, Montreal, Canada}
\icmlaffiliation{mit}{MIT Media Lab, Cambridge, MA, USA}

\icmlcorrespondingauthor{Prashanth Vijayaraghavan}{pralav@mit.edu}

\icmlkeywords{Machine Learning, ICML}

\vskip 0.3in
]



\printAffiliationsAndNotice{} 


\begin{abstract}
With growing role of social media in shaping public opinions and beliefs across the world, there has been an increased attention to identify and counter the problem of hate speech on social media. Hate speech on online spaces has serious manifestations, including social polarization and hate crimes. While prior works have proposed automated techniques to detect hate speech online, these techniques primarily fail to look beyond the textual content. Moreover, few attempts have been made to focus on the aspects of interpretability of such models given the social and legal implications of incorrect predictions. In this work, we propose a deep neural multi-modal model that can: (a) detect hate speech by effectively capturing the semantics of the text along with socio-cultural context in which a particular hate expression is made, and (b) provide interpretable insights into decisions of our model. By performing a thorough evaluation of different modeling techniques, we demonstrate that our model is able to outperform the existing state-of-the-art hate speech classification approaches. Finally, we show the importance of social and cultural context features towards unearthing clusters associated with different categories of hate. 
\end{abstract}

\section{Introduction}
\label{intro}
Recently, Social media has become a breeding ground for abuse in the form of misuse of private user details \cite{yang2014translating,kontaxis2011detecting}, hate speech and cyberbullying \cite{djuric2015hate,sood2012profanity,singh2017they}, and false or malicious information \cite{cheng2017anyone,aro2016cyberspace,weimann2010terror}. 
M{\"u}ller et al. \cite{muller2017fanning,muller2018making} investigated the link between social media, specifically Facebook and Twitter and hate crime in countries like United States and Germany. The results of these studies suggest that social media can act as a propagation mechanism between online hate speech and real-life violent crimes. Therefore, the need for combating real-life hate crimes also involves improving the health of public discourse on such online platforms. The primary step in this process requires detecting and tracking hate online. This is also critical for protecting democracy from motivated malignancy. In this work, we, therefore, focus on hate speech -- defined formally as any communication that disparages a person or a group on the basis of some characteristic such as race, color, ethnicity, gender, sexual orientation, nationality, religion, or other characteristic \cite{nockleby2000hate}.

A recent study \cite{grondahl2018all,hosseini2017deceiving} has exposed weaknesses in many machine learning detectors currently used to recognize and keep hate speech at bay.
In addition to problems in dealing with language related challenges, prior work ignores an important aspect of hate speech -- social and cultural dimension of hate speech. Hate speech covers a range of different type of expressions and it is critical to look at the context in which a particular expression is made: (a) What was the nature of expression? (b) Who made the expression? and (c) What was the socio-cultural context? Since language also evolves quickly, specifically among young users of social networks, it adds another layer of complexity \cite{nobata2016abusive}. Some communities tend to use benign words or phrases that have accepted hate speech meanings within their community and specific social context of usage. Therefore, we need to understand that hate speech may have strong socio-cultural implications \cite{raisi2016cyberbullying} that needs to be addressed. 

Most of the current methods are trained primarily using the textual features containing limited examples of hate content and the results of their evaluations suggest that it is really difficult to detect hateful content without any additional contextual information \cite{kwok2013locate}. In this work, we develop a hate speech classification model for tweets incorporating both social and cultural background of the user along with textual features. We further analyze the tweets to understand how the socio-political context helps us understand the categories of hate speech.


 


\section{Dataset}
\label{all_data}
One of the main challenges in building a robust hate speech detection model is the lack of a common benchmark dataset. Our dataset consists of tweets collected and aggregated from various different sources.

\subsection{Publicly available datasets}
There are different publicly available datasets usually obtained using keyword, hate lexicon or hashtag based filtering from Twitter stream over a period of time and then manually annotated. Table \ref{datasets} gives details of all datasets used in our work. 
It is found that distinguishing hate speech from non-hate offensive language is a challenging task, as presence of offensive words is not a necessary condition for hate speech and offensive language does not always express hate.

\begin{table*}[!htb]
  \centering
  \begin{tabular}{l|l}
    \toprule
    \textbf{Datasets} & \textbf{Details} \\
    \midrule
   \cite{founta2018large} & None: 53.8\%; Hate: 4.96\%; Abusive: 27.15\%; Spam: 14\%; Tweets: $\sim{100k}$;\\
    \cite{davidson2017automated} &  None: 16.8\%; Hate: 5.8\%; Offensive: 77.4\%; Tweets: $\sim{25k}$; \\
     \cite{park2017one} &  None: 68\%; Sexism: 20\%; Racism: 11\%;  Tweets: $\sim{18k}$; \\
    \cite{golbeck2017large} &    None: 74\%; Harassment: 26\%; Tweets: $\sim{21k}$;\\
    Our dataset &    None: 58.1\%; Hate: 16.6\%; Abusive: 25.3\% Tweets: $\sim{258k}$;\\
    \bottomrule
  \end{tabular}
  \caption{Summary of different datasets.}
  \label{datasets}
\end{table*}

\subsection{Data collection}
We further expand the existing dataset using data augmentation and distant supervision techniques. We collect additional data to add more capabilities to our modeling. We explain the steps involved in collecting these additional resources:

\paragraph{Tweets}
In order to expand our dataset, we perform an exploratory search on Twitter that consists of phrases containing:
(a) swear words combined with positive adjectives (eg. ``f**king awesome'', ``damn good'', ``bloody wonderful'')
(b) swear words combined with races, religions, sexual orientations or derogatory references to them. (e.g. ``f**king ragheads'', ``sh**ty muslims'').  
We sample hundred random tweets from a list of tweets collected using (a) and (b) and manually annotated them. With high precision, the tweets obtained using (a) were non-hateful, while those tweets collected using (b) comprised of hate messages. This is backed by the procedure followed by Golbeck et al.\cite{golbeck2017large,hatelingo} and subsequently manually annotated. We also get the hate code words as explained by Magu et al. \cite{magu2017detecting} and obtain those tweets containing such code words along with swear words and identity based derogatory lexicons. We curate few random samples of this aggregated data using human annotators and majority of the annotation conformed with the classes they were assigned to. We perform data augmentation by introducing some noise in the form of common misspellings into the data.

\begin{figure}[!htb]
\centering

\includegraphics[width=0.5\linewidth]{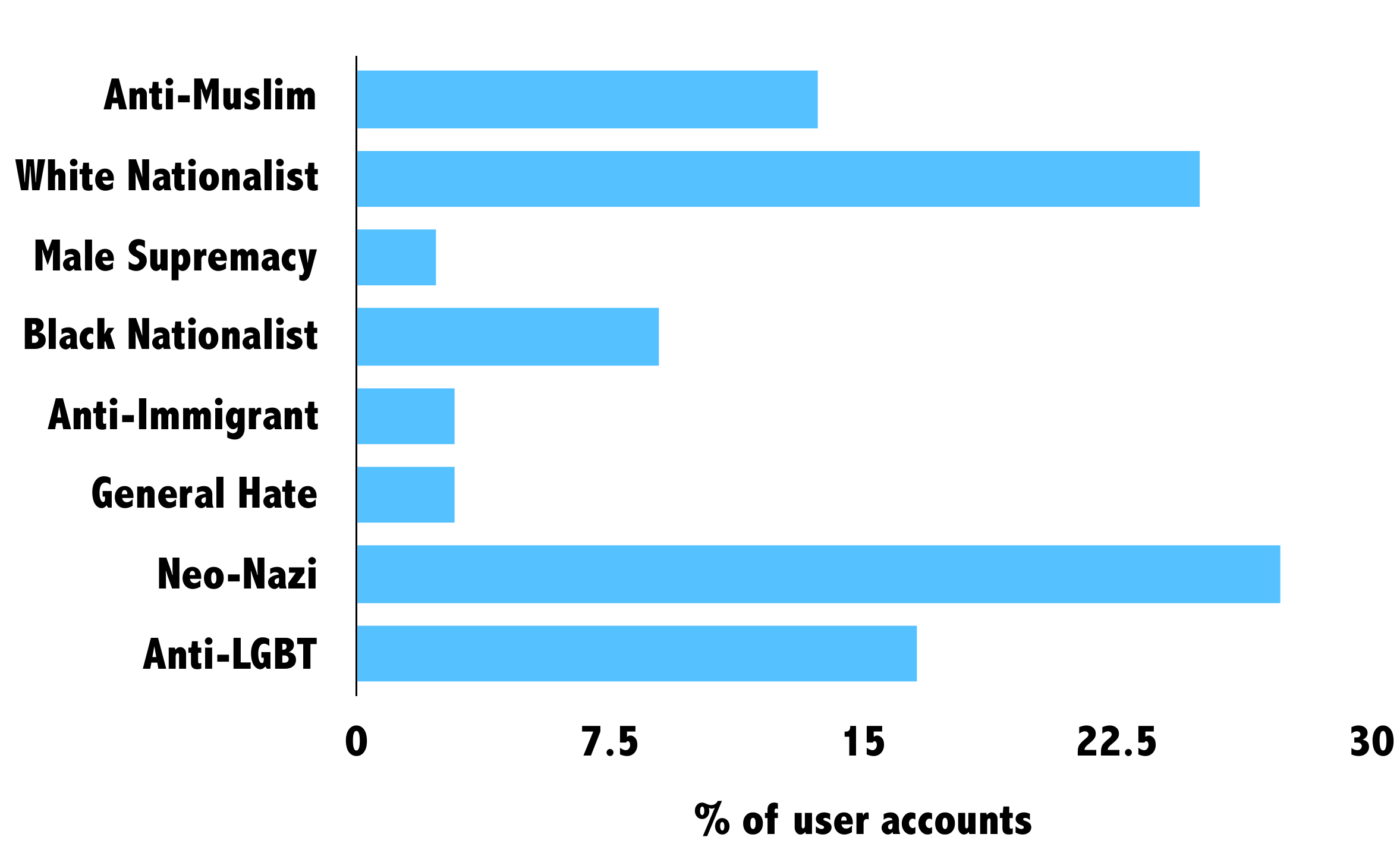}
\includegraphics[width=0.45\linewidth]{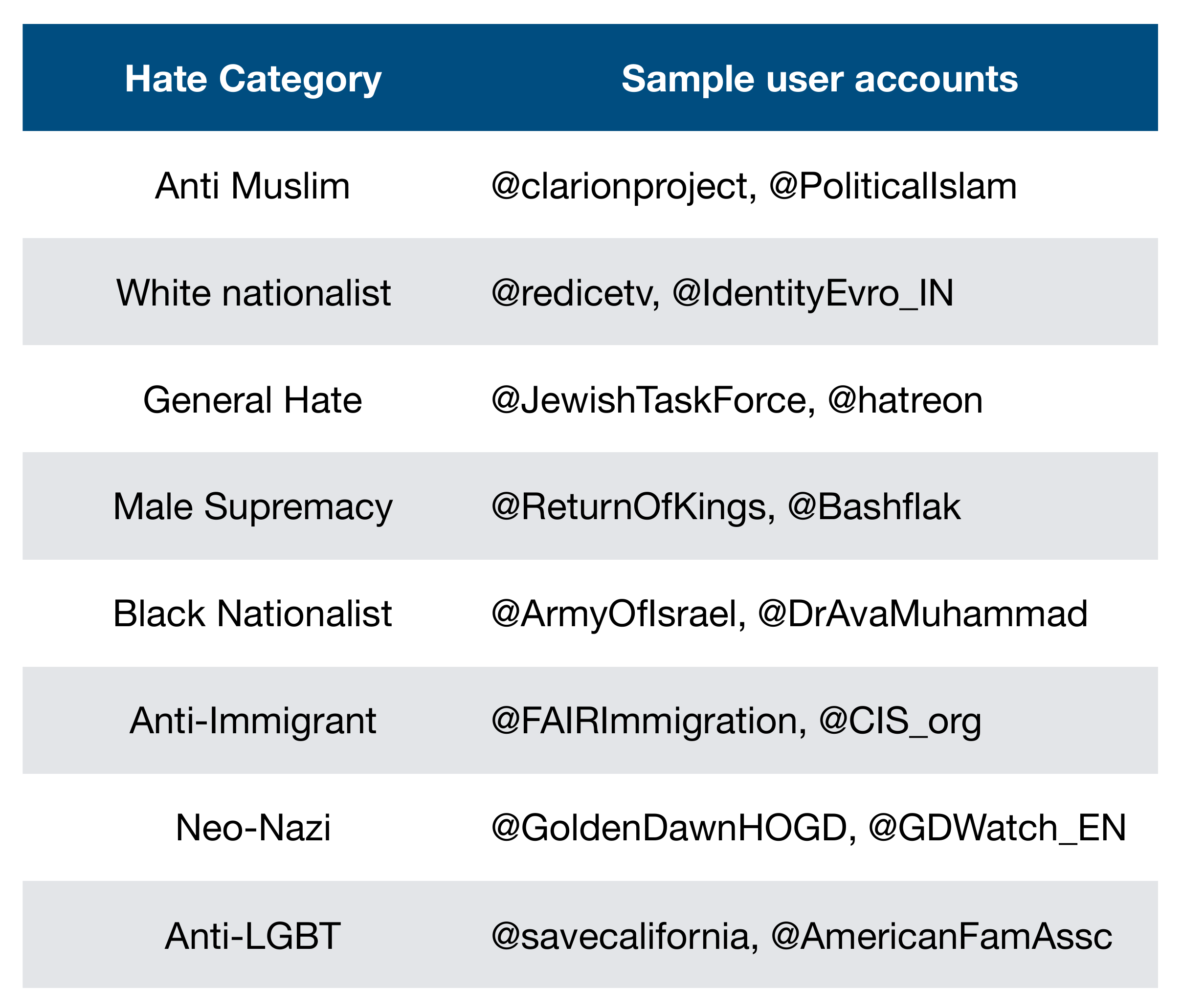}

\caption{Left: Percentage of seed user accounts belonging to a particular hate category. Right: Sample user accounts related to each hate category.}
\label{hate_group}
\end{figure}

\paragraph{Hate groups} Collecting information about extremist groups and their followers can be useful in determining hate communities on Twitter. In this direction, we gather the data collected by Southern Poverty Law Center (SPLC)\footnote{https://www.splcenter.org/}. SPLC has been monitoring activities of various extremist groups in United States like anti-immigrant, white nationalists, the neo-Nazi movement, antigovernment militias and others. The SPLC website contains information about prominent members of such groups and their websites. In an effort to map these extremist groups on twitter, we perform a user account look up on Twitter by filtering those users who have the extremist group's name as apart of their name or bio. We got $\sim{3k}$ user accounts containing such information and filtering for inactive accounts. We construct a directed graph $\mathcal{G}$ where each vertex is a user and edges represent their relationships (friends \& followers). We compute the page rank of this graph $\mathcal{G}$ and obtain the top $\sim{10k}$ accounts including the $\sim{3k}$ seed user accounts. Figure \ref{hate_group} shows the the various categories of hate groups and summary of the the data collected. we construct a graph $ G^h$ using all the authors' and ``mentioned'' users' follower and friends list in our dataset and perform an intersection with the $\sim{10k}$ hate community accounts.

Finally, we construct our dataset removing data with no user information. Table \ref{datasets} gives the statistics of the aggregated dataset. We believe that all these resources can help us build an efficient hate speech classifier, considering language, social and cultural context as a whole. 
\begin{figure}
\centering
\includegraphics[width=0.65\linewidth]{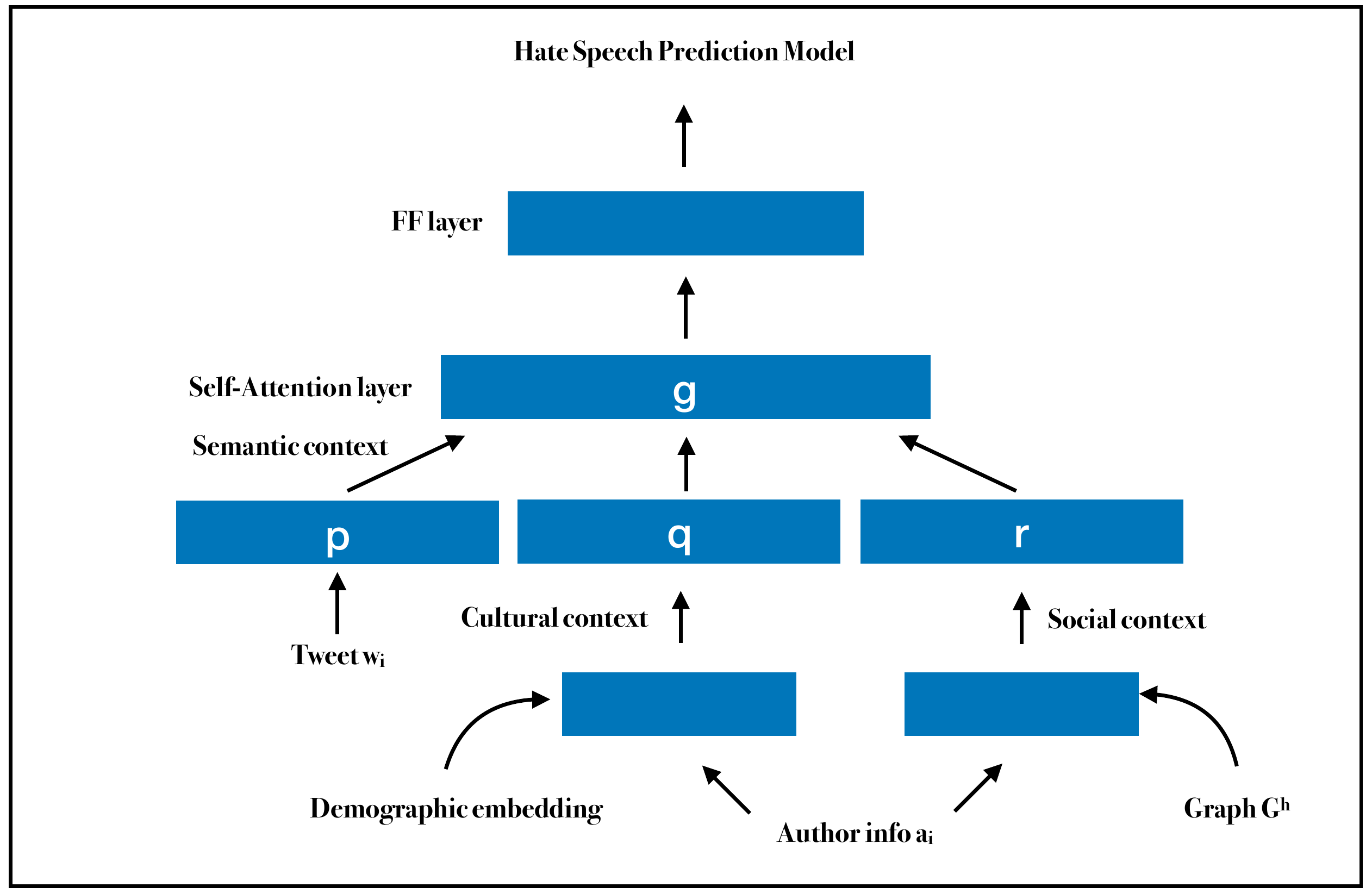}
\caption{Illustration of the model}
\label{hate_model}
\end{figure}

\section{Model}
\label{modell}
In this section, we describe our deep learning model that accounts for the multimodal nature of our data. The data includes tweets, author attributes and social structure represented by author's followers and friends. Let us denote our hate dataset as $D^{(H)}={(w_1, a_1),(w_2, a_2),....,(w_n, a_n,)}$, where each tuple in this set consists of the tweet text $w_i$, author information $a_i$ used to derive social and cultural context associated with the tweet. The dataset $D^{(H)}$ is divided into $D^{(H)}_{train}$, $D^{(H)}_{val}$ and $D^{(H)}_{test}$ for training, validation and testing purposes. 
Defining the input as $x_i=(w_i,a_i)$, we denote our model as:
\begin{equation}
    f_\theta(x_i) \approx g_\theta(P(w_i), Q(a_i), R(a_i))
\end{equation}
where $P,Q,R$ are neural architectures extracting semantic features from tweet text ($P$) and socio-cultural features($R,Q$) from author information; $g$ is the function that determines the fusion strategy. Figure \ref{hate_model} is an illustration of our model. 
\subsection{Extracting semantic features (p)}
 There are a number of recent works \cite{kim2014convolutional,kiros2015skip,conneau2017supervised} that have focused on encoding the text into useful continuous feature representations. Our choice of the text encoder is motivated by the need to encode character information along with the word information to handle noisy nature of tweets. Our encoder is a slight variant of Chen et al.\cite{chen2018combining}. This approach providing multiple level of granularity can be useful in order to handle rare or noisy words in the text. Given character embeddings $E^{(c)}=[e_1^{(c)}, e_2^{(c)},...e_{n'}^{(c)}]$ and word embeddings  $E^{(w)}=[e_1^{(w)}, e_2^{(w)},...e_{n}^{(w)}]$ of the input, starting ($p_t$) and ending ($q_t$) character positions at time step $t$,  we define inside character embeddings as: $E_I^{(c)}=[e_{p_t}^{(c)},..., e_{q_t}^{(c)}]$ and outside embeddings as: $E_O^{(c)}=[e_{1}^{(c)},...,e_{p_t-1}^{(c)}; e_{q_t+1}^{(c)},...,e_{n'}^{(c)}]$. First, we obtain the character-enhanced word representation $\overleftrightarrow{h_t}$ by combining the word information from $E^{(w)}$ with the character context vectors. Character context vectors are obtained by attending over inside and outside character embeddings. Next, we compute a summary vector $S$ over the hidden states $\overleftrightarrow{h_t}$ using a self-attention mechanism expressed as $Attn(\overleftrightarrow{H})$. The summary vector $S$ is the final textual feature representation. The attention mechanism helps us to recognize the most relevant text units that contribute towards hate speech prediction.

\subsection{Extracting cultural context features (q)}
 The demographic features act as a proxy for deriving the cultural context information required for our classification task. A work by Vijayaraghavan et al. \cite{vijayaraghavan2017twitter} developed a multimodal demographic classifier for Twitter users. By feeding author information $a_i$ to this model, we extract the penultimate layer demographic representations for all users in our dataset.
 \begin{figure}
\centering
\includegraphics[width=0.7\linewidth]{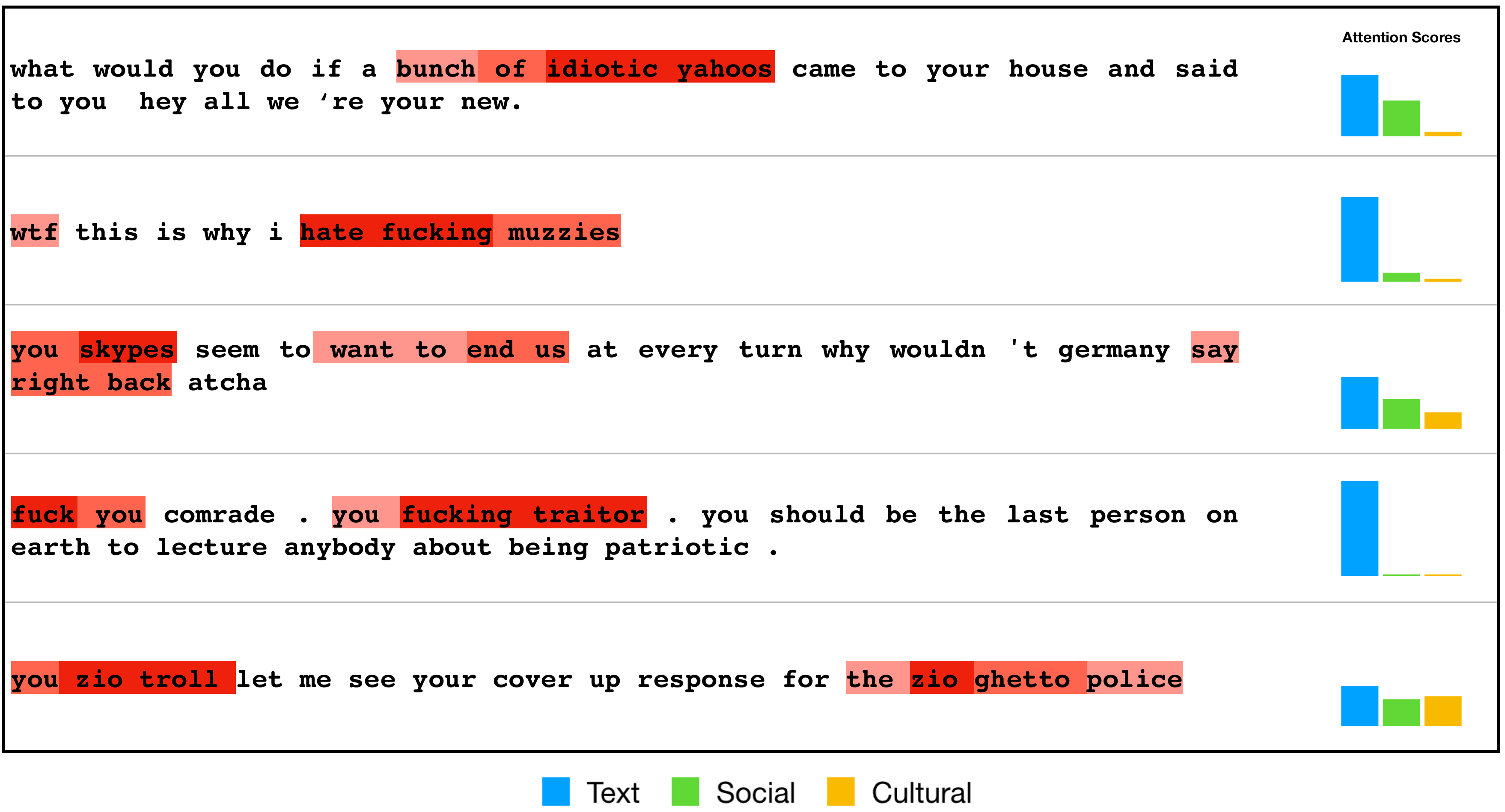}
\caption{Table shows examples with attention scores of textual, social and cultural context features. Text containing hate code words have higher attention scores (highlighted) for social and cultural context features.}

\label{attn}
\end{figure} 
 \subsection{Extracting social context features (r)}
 Social context is obtained using the graph $G^h$ explained in Section \ref{all_data}. We use an approach delineated by Vijayaraghavan et al. \cite{vijayaraghavan2017twitter}, we build a binary vector for each tweet's author $a_i$ based on the  $\sim{10k}$ hate community accounts. Each index of the vector represents one of those hate group accounts the value (1, 0) represents if the author is following the account or not. We use feed-forward layers to map this binary vector into a lower dimension vector. 
 
 \subsubsection{Combining textual and socio-cultural features (g)}
 We apply a late fusion strategy in our work. The multi-modal features are fed through a self-attention mechanism to understand the relevance of each of these different features towards the hate speech classification task. We perform a weighted sum of these feature vectors to obtain a final vector representation $R^h$. We use categorical cross-entropy as our objective to train our model.
 

\setlength{\abovedisplayskip}{0cm}
\small

\section{Evaluation}
We experimented with different models. Table \ref{results} shows the results of our evaluations using traditional and other baseline deep learning models. It is clear that deep learning models outperform the traditional models. Table \ref{results} also shows that social and cultural context improves the performance of our model significantly compared to purely state-of-the-art text-based models. 

\begin{table}
\small
\begin{center}

\begin{tabular}{c|c|c}
\toprule
\bf{Model} & \bf{F1 (Hate)} & \bf{F1 (Overall)}\\
\midrule 
	
    \multicolumn{3}{c}{Traditional Models: Text+Social+Demographic} \\ \midrule
    LR & 0.53 & 0.72  \\
    SVM  & 0.563 & 0.729 \\
    \midrule
    \multicolumn{3}{c}{DL Models: Text Only} \\ \midrule
    CNN-Char & 0.735 & 0.866  \\
    BiGRU+Char+Attn & \textbf{0.744} & \textbf{0.864}\\
    CNN-Word & 0.658 & 0.788  \\
    BiGRU+Attn & 0.683 & 0.801  \\
    BiLSTM-2DCNN & 0.661 & 0.795 \\
    \midrule 
    \multicolumn{3}{c}{DL Models: Text+SC } \\ \midrule
    CNN-Char+FF & 0.760 &  0.879  \\
    BiGRU+Char+Attn+FF & \textbf{0.784} & 
    \textbf{0.90}  \\
\bottomrule
\end{tabular}
\end{center}
\caption{\label{results}Results of our experiments. }

\end{table}

\subsubsection{Identifying categories of hate speech}
We verify if the embeddings learned from the model can help identify different hate categories. To achieve that, we compute the final representation $R^h$ before the final feed-forward layer  for every tweet classified as hate speech by our model. This representation $R^h$ is our hate embedding. We use a simple agglomerative clustering method on our hate embeddings with a parameter $k=5$ for the number of clusters. We manually label these clusters using the top words ranked based on attention scores. Table \ref{attn_words} shows the top 5 words associated with the clusters and manual hate category assigned to them. 

\begin{table}[h!]
\small
  \begin{center}
  \begin{tabular}{|p{0.25\textwidth}|p{0.15\textwidth}|}
    \hline
     \textbf{Top 5 words} & \textbf{Hate Category}  \\
    \hline
      jihadi, muzzie, terrorist, \#stopislam, \#banmuslim & Anti-Islam \\\hline
     n**ga, n**ger, \#whitepower, ghetto \#14words & Anti-Black  \\\hline
      \#buildthewall, illegals, \#noamnesty, \#illegalaliens, \#anchorbabies & Anti-Immigrant \\\hline
      f**k, c**t, hate, b**ch, a****le & General Hate\\\hline
       \#antisemitism, \#antisemites, nazi, satan, neonazi   & Anti-semitic \\
    \hline
  \end{tabular}
  \end{center}
  \caption{Top 5 words associated with different clusters and manually assigned hate category.}
  \label{attn_words}
\end{table}

 We evaluate the relevance of social and cultural context features by comparing the clusters produced by embeddings obtained from two models: Text-Only Model (BiGRU+Char+Attn) and Text+SC Model (BiGRU+Char+Attn+FF). We measure the ability of these models to recover the hate categories with our embeddings through a purity score. We measure the amount of overlap between two clusterings as: $Purity = \frac{1}{N}\sum_i max_j |g_i \cap c_j|$, where $N$ is the total number of input data points, $g_i$ is the $i$-ith ground truth hate category, and $c_j$ is the $j$-th predicted cluster.

 Since we do not have the ground truth hate category for our tweets, we sampled 100 random tweets from every cluster obtained from both the models and obtained human judgments. Workers were shown the original tweet and asked the hate category they fall under. They were also allowed to choose 'None of the above' as an option. Once we obtain the ground truths using these human annotations, we computed the purity scores. The results are shown in Table \ref{tab_purity}. The results indicate that the model that uses social and cultural context is able to produce better clusters having more overlap with ground truth hate categories compared to the text only model. The performance can be attributed to the Hate graph $G^h$ containing inherent information of various categories of hate groups. 

\begin{table}[h!]
\small
\begin{center}
\begin{tabular}{|c|c|}
\hline $Model$ & $Purity$ $Score$   \\ \hline
Text Only &  \textbf{0.52} \\\hline
Text+SC &  \textbf{0.76} \\
\hline
\end{tabular}
\end{center}
\caption{Cluster purity scores}
\label{tab_purity}
\normalsize
\end{table}

\subsection{Interpretability}
 In order to be able to draw insights on the decisions of the model, we qualitatively look at the data where the social context information is used for prediction. In Section \ref{modell}, we explained the need for an attention layer that fuses both the semantic text features and social and demographic features. In order to assess the importance of social and cultural context for hate speech classification, we identify examples in test set that has a high attention score for social context vector. We visualize by highlighting words in the text and constructing bar graphs that indicate the relevance of each of the features: textual, social, cultural. We compared the results from attention scores with perturbations-based approach \cite{ribeiro2016should, guidotti2018survey} for interpretability. Since they were strikingly similar, we show the results based on attention scores only. Figure \ref{attn} shows few examples where that is the case. Interestingly, these are tweets containing code words like ``skypes'', ``yahoos'', ``zio'', ``zog'', etc. attacking particular group of people. These kinds of code words are widespread among the hate groups (explained in Section \ref{all_data}). Though our dataset contains only few instances of such words in the presence of swear words, the model is still able to understand these code words and tag them as hateful content which otherwise could just pass on as being abusive but not hateful (though it can be debatable).

\section{Conclusion}
In this work, we developed a comprehensive model to automatically detect hateful content. Motivated by the need to incorporate socio-cultural information for hate speech detection, we adopt different feature extraction strategies for different modalities of data: text, demographic information and social graph. We build a multi-modal classification method with late-fusion of different modalities of data. We derive important insights about our model and its ability to understand hate speech code words and cluster into different categories of hate speech.



\bibliography{icml19}
\bibliographystyle{icml2019}





\end{document}